\documentclass{article}

\usepackage{PRIMEarxiv}
\usepackage{url,hyperref,lineno,microtype,subcaption, amsmath, amssymb}
\usepackage[onehalfspacing]{setspace}
\usepackage{cite}
\usepackage{amsmath,amssymb,amsfonts}
\usepackage{algorithmic}
\usepackage{algorithm}
\usepackage{graphicx}
\usepackage{textcomp}
\usepackage{url}      
\usepackage{upgreek}       
\usepackage[dvipsnames]{xcolor}     
\usepackage{pifont}
\usepackage{footnote}
\usepackage{xcolor}
\usepackage[utf8]{inputenc} 
\usepackage[T1]{fontenc}    
\usepackage{hyperref}       
\usepackage{nicefrac}       
\usepackage{microtype}      
\usepackage{lipsum}
\usepackage{fancyhdr}       
\usepackage{graphicx}       
\graphicspath{{media/}}     
\usepackage{tabularx,booktabs}

\DeclareMathOperator*{\argmin}{arg\,min}
\title{Low-earth Satellite Orbit Determination Using Deep Convolutional Networks with Satellite Imagery
}

\author{
  Rohit Khorana \\
  Saint Francis High School \\
  Mountain View, CA\\
  \texttt{rohit.khorana.email@gmail.com}
}

\begin{document}
\maketitle

\begin{abstract}
Given the critical roles that satellites play in national defense, public safety, and worldwide communications, finding ways to determine satellite trajectories is a crucially important task for improved space situational awareness. However, it is increasingly common for satellites to lose connection to the ground stations with which they communicate due to signal interruptions from the Earth's ionosphere and magnetosphere, among other interferences. In this work, we propose utilizing a computer vision based approach that relies on images of the Earth taken by the satellite in real-time to predict its orbit upon losing contact with ground stations. In contrast with other works, we train neural networks on an image-based dataset and show that the neural networks outperform the de facto standard in orbit determination (the Kalman filter) in the scenario where the satellite has lost connection with its ground-based station. Moreover, our approach does not require {\it a priori} knowledge of the satellite's state and it takes into account the external factors influencing the satellite's motion using images taken in real-time.
\end{abstract}


\section{Introduction}
It is quite common for satellites to lose connection with their ground-based stations \cite{yao2017gps, seo2011availability}. The \emph{Swarm} satellites lost GPS signal many times during their time in space. The loss of signal was attributed to interference from equatorial plasma irregularities (EPIs) \cite{xiong2016swarm}. It has also been shown that ionospheric irregularities can interfere with satellites' GPS connection and lead to signal fading \cite{carrano2010temporal}. This ionospheric scintillation is attributed to electron density irregularities in the ionosphere \cite{seo2011availability}. satellites in orbit experience many other unpredictable disturbances such as solar radiation, atmospheric phase screening, and atmospheric drag, all of which interfere with communications between satellites and have resulted in a loss of connection with ground stations \cite{ODCurtis}. Accurate monitoring in such scenarios is required, as inaccurate prediction of satellite orbits could lead to satellite collisions and an increase in space debris \cite{chen2023research}. Over the past 50 years, orbit determination solutions have been developed and refined for this very purpose, and the current methods achieve notable accuracy.

Satellite orbit determination is the computational process of determining the satellite's state, or ephemeris, as a function of time using sets of measurements collected onboard the satellite or by ground-based tracking stations \cite{schutz2004, ODfifty}. 
The classical method for orbit determination relies on a Kalman filter. However, the orbital data required for the Kalman filter to work is provided by systems of well-connected active radar calibrators (ARCs) that provide range and range-rate measurements. ARCs send pulses regularly from antennas and store the resulting  echoes in a two-dimensional (2D) matrix, which is referred to as the raw data. Then, through the use of an synthetic aperture radar (SAR) processor, a focused image is obtained after two compressions have been performed on the received raw data \cite{GEOSAR, introSAR}. The first compression is performed along the range direction, using a matched filter, and the second is performed along the azimuth direction, with the signal being compressed by an SAR algorithm. The primary issue with the use of SAR, and ARC, however, is that the lateral resolution degrades with increasing operating distance \cite{GEOSAR}, and other factors such as frequency shift can exacerbate the inaccuracies even further \cite{SARARCIEEE}.

After the data provided by the ARC has been obtained, the next step in determining the orbit of a satellite is to apply initial orbit determination (IOD) techniques. The primary technique for IOD is the Gibbs method, which takes as input three geocentric position vectors taken at three different times and then calculates the state vector at the second timestamp \cite{gibbs}. It should be noted, however, that the Gibbs method requires all input position vectors to be co-planar \cite{gibbs}. This preliminary orbit can then be refined through the use of differential correction techniques, the most common being Kalman filtering. The Kalman filtering approach uses Global navigation satellite system (GNSS) measurements in tandem with an equation describing the motion of the system \cite{KFScirp}. Although this method is quite advanced, the primary issue with this approach is the reliance on GNSS measurements, which will not be accessible during the period where the satellite has lost its connection with ground based stations. Thus, the process of accurately determining the satellite's orbit in such a scenario is increasingly difficult.

In this work, we develop a novel computer vision approach for satellite orbit determination that represents a viable solution in the increasingly common scenario where the satellite has lost connection with its ground based stations. We utilize a real-time image, taken by the satellite, to predict the satellites' orbit using hardware either already present on the satellite or hardware which can be easily added (camera, RISC-V processor for IoT applications). The methodology described in this paper represents a high-accuracy machine learning approach for full orbit determination without using {\it a priori} information of the satellite's ephemeris. It is important to note we are the first to use machine learning for the whole process of satellite orbit determination \cite{haidar2022hybrid}.

\section{Related Work}

\subsection{Deep Learning in Satellite/RSO Orbit Prediction}
Many studies have examined the application of machine learning (ML) to the prediction of LEO satellites' orbits. Many authors find machine learning methods attractive due to their ability to capture the complex, non-linear dynamics of LEO satellites at a low computational cost, relative to other orbit propagation methods \cite{mortlock2021assessing, haidar2022hybrid, peng2018artificial}. In \cite{mortlock2021assessing}, a time-delay neural network (TDNN) is trained based on data from two Orbcomm satellites, which broadcast their signal to a ground-based navigating vehicle. The vehicle decodes this signal using Doppler measurements to obtain the satellite's position. 
A hybrid analytical-machine learning approach whereby a stationary receiver tracks a LEO satellite using an Extended Kalman Filter (EKF) initialized with SGP4-propagated TLE is proposed in \cite{haidar2022hybrid}. These measurements are used to train a nonlinear autoregressive with exogenous inputs (NARX) neural network, which can then predict the future orbit of the satellite. Li et al. perform error propagation by utilizing a ML model to fit the historical error of orbit predictions for space debris in \cite{li2020machine}. This ML model can be used to predict the error of future physics-based orbit predictions and correct said predictions. In \cite{peng2017limits}, a support vector machine (SVM) is trained to estimate errors associated with orbit prediction estimates and generate a ML-modified state, supposedly closer to the satellite's true state. The same authors applied ML in a similar fashion in \cite{peng2018artificial}, except an Artificial Neural Network (ANN) is used to generate the ML-modified state instead of an SVM. 

A follow-up simulation study was conducted by the authors in \cite{peng2019comparative}, where they show that ANN has the best orbit prediction capabilities, when compared to that of the SVM and Gaussian processes. In 2020, the authors used an SVM to map low-accuracy TLE catalog to high-accuracy International Laser Ranging Service (ILRS) catalog \cite{peng2020machine}. In their most recent endeavor, those authors take a fusion approach, where they apply machine learning methods, trained on historical error data, in tandem with the conventional framework, the EKF \cite{peng2021fusion}. Note that using an EKF in a situation where the satellite has lost connection with its ground-based station significantly hinders the proposed fusion system. In \cite{salleh2019enhancing}, a TDNN and Long Short-term Memory (LSTM) were used to enhance the Simplified General Perturbations-4 (SGP4) model, a common orbit propagation model. In \cite{kozhaya2021comparison}, an ephemeris propagation framework and error propagation framework are presented. In the ephemeris propagation framework, a neural network is trained to predict an Orbcomm satellite's future orbit, after being trained on historical HPOP data and decoded Orbcomm satellite ephemeris messages. The authors note that the time needed for inference is a limiting factor of the framework \cite{kozhaya2021comparison}. In the error propagation framework, a NARX neural network is trained to map SGP4 propagated state vectors to the more-accurate HPOP propagated state vectors, essentially using a NARX neural network to correct SGP4 propagated state vectors to map to HPOP state predictions. The authors note that the error propagation framework outperforms the ephemeris propagation framework. It should also be noted that SGP4 is a less-accurate, but faster, state propagator than HPOP. 

\subsection{Classification of Deep Learning in Satellite/RSO Orbit Prediction}
All of the studies in section 2.2 of this paper present techniques which can be classified as belonging to one of three approaches:
\begin{itemize}
    \item [(1)] Analytical + Machine Learning Ephemeris Prediction [AML-EP]
    \item [(2)] Analytical + Machine Learning Error Prediction [AML-ERP]
    \item [(3)] Refined Kalman Filter Ephemeris Prediction [RKF-EP]
\end{itemize}

\subsubsection{Analytical + Machine Learning Ephemeris Prediction}
In the AML-EP approach, analytical methods use LEO satellite signal or GPS signal to track the satellite and estimate the satellite's ephemeris; this information is used by machine learning methods to predict the satellite's orbit.
The AML-EP methods can be divided into two sub-categories:
\begin{itemize} 
    \item[(Type I)] The estimated ephemeris is used as the ground-truth for the LEO satellite trajectories during the training of the neural networks. After training, the neural network will be able to propagate the state for that specific satellite. Studies \cite{haidar2022hybrid} and \cite{mortlock2021assessing} resemble this framework.
    \item[(Type II)] The estimated ephemeris is propagated by a numerical or computational propagator, and the state vectors generated are used to train the neural network, to model said propagator. HPOP was used as the numerical propagator in \cite{kozhaya2021comparison}. After training, the neural network can take historical state data of a satellite as input, and accordingly propagate the satellite's ephemeris. 
\end{itemize}

The limitation of Type I approaches is that the analytical methods used require a LEO satellite signal or GPS signal to obtain the measurements of the satellite. Thus, the neural network has to be retrained on new measurements for each LEO satellite's orbit it wants to predict. Our approach avoids this limitation as our CNN just takes in an image as input and only needs to be trained one time. One can train the model, save the weights and use those weights for orbit predictions for any satellite constellation. As a result, we do not need to constantly retrain a neural network for each satellite whose orbit we wish to predict.

The limitation of Type II approaches are that the neural network takes a longer time for inference if they are modeling a high-fidelity propagator, such as HPOP. If neural networks are used to model lower-fidelity propagators instead, like SGP4, the accuracy achieved would be substantially lower. This trade-off is noted in \cite{haidar2022hybrid}. Regardless of the numerical/analytical propagator, the neural network relies upon {\it a priori} knowledge of the satellite's state to predict the ephemeris. Our approach is therefore more versatile, as we do not require any prior knowledge of the satellite's orbit. Thus, we can predict the state of a satellite even if the historical dataset for the analytical/computational propagator does not exist yet. Additionally, our approach does not pose any significant trade-offs between accuracy and model inference time. 

\subsubsection{Analytical + Machine Learning Error Prediction}
In the AML-ERP approach, analytical methods are used to obtain information about the LEO RSO's ephemeris, and machine learning methods are used to predict errors associated with various physics-based orbit predictions. The information about the object's orbit is used to generate a historical dataset, for said object, mapping physics-based orbit predictions to their corresponding errors. Machine learning methods, such as Boosted Trees, SVMs, ANNs, and Gaussian Processes, among others, are trained to learn the relationship between orbit predictions and their errors. So, after training, the methods can predict the errors of future physics-based orbit predictions of said object, as in \cite{li2020machine, peng2017limits, peng2018artificial, peng2019comparative, peng2020machine, salleh2019enhancing, kozhaya2021comparison}.

A limitation of this approach is that there needs to be historical dataset constructed for the particular RSO. As a result, the machine learning methods which are trained on this dataset will not be able to generalize, and predict errors of physics-based approaches for other RSO's. Further, the approach relies on physics-based models for the baseline estimate, like SGP4 among other computational/numerical propagators. These propagators use incomplete perturbation force models to make the initial orbit prediction, and these predictions degrade in accuracy as time goes on \cite{li2020machine}. 

Thus, even though machine learning is employed to account for inaccuracies made by these dynamic force models, it has been shown that the randomness introduced during measurement of the system cannot be completely removed; in fact, as measurement errors enlarge, so do the errors of machine learning methods' predictions \cite{peng2017limits}. Therefore, if the assumed model is inaccurate, machine learning corrections will not be able to completely fix the error 

Our approach resolves these issues because our CNNs are able to generalize to any satellite (or RSO) provided it has the ability to take an image. Furthermore, we do not rely on any measurements taken by ground stations or onboard receivers, which may contribute to noise, as input to our CNN. In fact we specifically design our approach for the event of losing connection with ground based stations. Additionally, we do not rely upon any state propagators, such as SGP4, thereby avoiding assuming an incomplete force model.

\subsubsection{Refined Kalman Filter Ephemeris Prediction}
In the RKF-EP approach, Peng et al. built upon their previous work, where they developed a framework which employed machine learning methods to correct physics-based orbit predictions \cite{peng2021fusion}. In their work, they fuse this machine learning approach with the classical method for orbit prediction, the EKF, by generating a machine learning-modified state from an initial estimation provided by the EKF. Although this methodology achieves notable accuracy, the performance is significantly worse in the scenario where a satellite has lost connection to its ground-based station.

This degraded accuracy lies in the fact that the EKF, without a GPS signal to refine its estimates, reduces to a standard state propagator. Thus, in the scenario analyzed in this study, the approach taken in \cite{peng2021fusion} is equivalent to that of the other AML-ERP approaches.

Our CNN-based approach addresses the limitations of the AML-ERP approach, due to our lack of reliance on ground-based stations, physics-based state propagators, or historical ephemeris data: we show high-accuracy satellite orbit predictions, in the common case of satellites losing connection, without the need for {\it a priori} information.

\begin{table*}
\caption{We categorize each work}
\label{table1}
\begin{tabularx}{\textwidth}{@{}lc@{}}
\toprule
Study & Classification \\
\midrule
Mortlock 2021 \cite{mortlock2021assessing} & AML-EP\\
Haidar 2022 ~~~~\cite{haidar2022hybrid} & AML-EP\\
Li 2020 ~~~~~~~~~~~\cite{li2020machine} & AML-ERP\\
Peng 2017 ~~~~~~~\cite{peng2017limits} & AML-ERP\\
Peng 2018 ~~~~~~~\cite{peng2018artificial} & AML-ERP\\
Peng 2019 ~~~~~~~\cite{peng2019comparative} & AML-ERP\\
Peng 2020 ~~~~~~~\cite{peng2020machine} & AML-ERP\\
Peng 2021 ~~~~~~~\cite{peng2021fusion} & RKF-EP\\
Salleh 2019 ~~~~~\cite{salleh2019enhancing} & AML-ERP\\
Kozhaya 2021 ~\cite{kozhaya2021comparison} & AML-EP and AML-ERP\\
\bottomrule
\end{tabularx}
\end{table*}

\newpage

\section{Methods}
In our setup we first must construct a dataset $\mathcal{D} := \{ (x_{i}, y_{i})\}_{i=1}^{N}$ for $N \in \mathbb{R}$, where $x_{i} \in \mathbb{R}^{a \times b \times 3}$ and $y_{i} \in \mathbb{R}^{3k \times 1}$. Each $x_{i}$ is an RGB image of the earth's surface and each $y_{i}$ is the corresponding ECEF vector which gives the position and orbit of the satellite.\\

\subsection{Dataset Construction}
We construct our dataset by collecting images and their corresponding metadata provided by Landsat 7 and Landsat 8, both satellites in low earth orbit. It should be noted that the format of the data is precisely the same regardless of whether the image and metadata came from Landsat 7 or Landsat 8. Due to storage limitations we only collect datapoints from certain bands (bands 3 and 4). To ensure diversity in the dataset we sample images uniformly at random from every major landmass, continent and ocean on Earth. We end up with approximately $10,000$ datapoints and after processing split the dataset into approximately $70 \%$ training, $20 \%$ validation and $10 \%$ test.

\subsubsection{Dataset Rows and Inputs}
It should be noted that one row from the dataset before any processing contains the following information:
\begin{minipage}{\textwidth}
\begin{savenotes}
\begin{center}
\begin{tabular}{|c | c |  c | c | c | c | c|} 
 \hline
 Image & Ephemeris Year & Ephemeris Day & Ephemeris Time & ECEF X & ECEF Y & ECEF Z \\
 \hline\hline
 $I \in \mathbb{R}^{a \times b \times 3}$ & $y \in \mathbb{R}$ & $d \in \mathbb{R}$ & $t \in \mathbb{R}^{k \times 1}$ & $x \in \mathbb{R}^{k \times 1}$  & $y \in \mathbb{R}^{k \times 1}$ & $z \in \mathbb{R}^{k \times 1}$\\  [1ex] 
 \hline
\end{tabular}
\end{center}
\vspace{1em}
\end{savenotes}
\end{minipage}
\\
This information is processed in two separate ways as the Neural Networks and Kalman filters rely on different columns from the same dataset to make predictions, since they are different methods.
\subsubsection{Neural Network Dataset}
The Neural Networks operate only on a subset of these columns namely the Image and ECEF X, ECEF Y, ECEF Z columns. Let $i$ denote the $i$-th row of the original dataset. Then the ECEF X, ECEF Y, ECEF Z columns are concatenated into one vector $V$ such that $V_{i} := [X_{i}/10000, Y_{i}/10000, Z_{i}/10000]$ where the division is applied elementwise. We apply a transformation to all of the images such that they all have the same shape of $256 \times 256 \times 3$.
Therefore the Neural Networks are trained on a dataset $\mathcal{D}_{nn} := \{(x_{i}, y_{i})\}_{i=1}^{N}$ where $x_{i} = I_{i} \in \mathbb{R}^{256 \times 256 \times 3}$ and $y_{i} = V_{i}$. This pre-processing does drop the Year, Day and Time information from consideration. One could perform feature engineering utilizing the Year, Day and Time information to likely improve the performance of the Neural Networks, however we knowingly choose not to do so. It should be noted that these rows are processed in order so that row $i$ from dataset $\mathcal{D}$ and row $i$ from dataset $\mathcal{D}_{nn}$ correspond to the same original datapoint. However, there is not necessarily a relationship between row $i$ and row $i+1$ in dataset $\mathcal{D}$ as the rows were shuffled before any model specific processing. 
\subsubsection{Kalman filtering Dataset}
The Kalman filters alternatively do not rely on image information to make predictions. Therefore the Image information is dropped. The Year, Day and Time information is converted to timestamp format. We additionally include the start time and end time as separate columns. Note that the Ephemeris Time column is actually a vector $t$ denoting a fixed range of times each spaced exactly $1$ second apart. In essence $t = [t_{0}, t_{1}, \ldots, t_{k-1}]$ where $t_{j+1} = t_{j} + 1$ for any $j \in [0, k-2]$. Then for every row in $D$ the ECEF X, ECEF Y, ECEF Z, and datetime information are converted to the ECI coordinate system to produce vectors $R1,~R2,~R3$. Therefore the Kalman filters are provided dataset $\mathcal{D}_{kf} := \{(R1_{i},R2_{i},R3_{i},t_{0},t_{k-1},timestamp)\}_{i=1}^{N} $. It should be noted that these rows are processed in order. Therefore row $i$ from dataset $\mathcal{D}$ and row $i$ from dataset $\mathcal{D}_{kf}$ correspond to the same original datapoint. However, there is not necessarily a relationship between row $i$ and row $i+1$ in dataset $\mathcal{D}$ as the rows were shuffled before any model specific processing. 

\subsubsection{The Neural Network Optimization perspective}

Let $\mathcal{F}$ denote a discrete set of neural networks. 
Then in essence $\forall~f_{\theta} \in \mathcal{F}$ and loss function $\ell(y_i, f_{\theta}(x_i))$ we wish to solve for optimal parameters $\hat{\theta} = \argmin_{\theta} \frac{1}{n} \sum_{i=1}^{n} \ell_{train}(y_{i}, f_{\theta}(x_i))$. The training loss $\ell_{train} = RMSE(y,\hat{y}) = \sqrt{\frac{\sum_{i=1}^{n} (\hat{y_i} - y_{i})^{2} }{n}}$. The test loss $\ell_{test}$ is RMSE. The optimizer utilized is SGD.\\

\subsection{Neural Network architectures and implementation}
We have tested various neural nets and have observed consistent phenomena. We  compare the performance of these neural networks with the Gibbs method and Kalman filtering. We discuss each of the model architectures and our implementation. 

\subsubsection{ResNet-101 and ResNet-50}

ResNet-50 is a type of residual convolutional neural network (CNN) with 50 layers. 
ResNet-101 is a type of residual CNN developed by He et al. \cite{ResNet}. Its architecture is identical to that of ResNet-50 except for the addition of three more layer blocks, leading to a total of 101 layers. 
Central to the architecture of the ResNet model is the notion of skip connections and identity mapping. In essence, given input $x$ and desired underlying mapping $H(x)$, He et al. \cite{ResNet} let $F(x):= H(x)-x$ and recast the initial mapping to $F(x)+x$. They argued that it is easier to optimize the residual mapping than the original un-referenced mapping. This formulation $F(x)+x$ is realized through skip connections, which skip one or more layers. We implement ResNet-50 using the ImageNet weights simply by importing both the model and the weights. We feed the output of  ResNet-50 through an additional GlobalAveragePooling2D layer, two dense layers with ReLU activations, and a final dense layer with linear activation. The loss function we seek to minimize is the RMSE, and the optimizer we use is stochastic gradient descent. We train the model for a total of 200 epochs and evaluate its performance on the test set. We do the same for ResNet-101.

\subsubsection{AlexNet}
AlexNet was devised by Krizhevsky et al.  \cite{AlexNet}. The model consists of eight layers with weights, namely, five convolutional layers and three fully connected layers. The first convolutional layer filters the image with a stride of four pixels, and the result is fed to a max-pooling layer. The second convolutional layer receives the output from the max-pooling layer and filters it with 256 kernels, and the remaining convolutional layers are connected without pooling or normalization layers in between \cite{AlexNet}. We implement the model as per the original paper \cite{AlexNet}, followed by a final dense layer with linear activation.
\subsection{Kalman filtering and Gibbs method}
\subsubsection{General Kalman filter} 
\paragraph{Description of Kalman filtering} 
The basic Kalman filter takes as inputs the initial state vector, the initial state error, the covariance of the process noise, the covariance of the observation noise, and measurements taken from sensors \cite{keil}. The Kalman filter is a recursive filter with two phases: prediction and update. In the prediction phase, the Kalman filter estimates the state at a later time, using the state transition matrix, which is derived from the Taylor series of the state at a certain time $t$ \cite{keil}. In addition, a new covariance will be produced, to approximate the uncertainty of the said prediction. In the update phase, a measurement of the state is taken via sensors. However, this measurement comes with some error, and the covariance of this measurement relative to that of the prediction is used to calculate the Kalman gain \cite{KFmath}. The Kalman gain represents the scaling factor, and it determines the relative impact of the sensor's measurement and predicted state on the updated state. 
\paragraph{Mathematical underpinnings} 
Mathematically, Kalman filtering is based on linear dynamical systems discretized in the time domain. To use the Kalman filter to estimate the internal state of a process given only a sequence of noisy observations for each time step, we specify the given time domain $T$ and $\forall t_{i} \in T$, the state transition model $F_{t_{i}}$, the observation model $H_{t_{i}}$, the covariance of the process noise $Q_{t_{i}}$,  the covariance of the observation noise $R_{t_{i}}$, and the control vector $u_{t_{i}}$. The Kalman filter supposes that the state at time $t_{i}$ is dependent upon the state at time $t_{i-1}$ according to $x_{k} = F_{t_{i}} x_{t_{i}-1} + B_{t_{i}}u_{t_{i}} + w_{t_{i}}$, with process noise $w_{t_{i}} \sim \mathcal{N}(0,Q_{t_{i}})$. At time $t_{i}$, an observation or measurement $z_{t_{i}}$ of the true state $x_{t_{i}}$ is made according to $z_{t_{i}} = H_{t_{i}}x_{t_{i}} + v_{t_{i}}$, where $H_{t_{i}}$ is the observation model and $v_{t_{i}}$ is the observation noise. 

\subsubsection{Kalman filter for orbit determination}

\paragraph{Extended Kalman filtering: current techniques for orbit determination}
Current approaches to the determination of satellite orbits generally use extended Kalman filtering, which is one of the most widely used estimators for nonlinear problems like orbit determination \cite{EKF}. The extended Kalman filter differs from the standard Kalman filter in that it first linearizes the problem at hand and then applies the linear Kalman filter to the resulting linear system \cite{EKF}.
The extended Kalman filter constitutes a state-of-the-art estimation algorithm for orbit determination \cite{KFgoldstandard}, or, more specifically, for predicting the future state vector of a satellite. In this context, continuous measurements are taken by GPS units, and so the extended Kalman filter's estimations are repeatedly refined. In other words, at every time step, the satellite position must be used to recalculate the state matrix and state transition matrix. Note that this algorithm is implemented in the discrete time domain. 
We let $x_{t} = f(x_{t-1}, u_{t}) + w_{t}$ and $z_{t} = h(x_{t}) + v_{t},$ where $u_{t}$ is the control vector, $w_{t} \sim \mathcal{N}(0,Q_{t})$ and $v_{t}\sim \mathcal{N}(0,Q_{t})$ are the process and observation noises, both of which are assumed to be zero-mean multivariate Gaussian with covariance matrices $Q_{t}$ and $R_{t}$, $f$ is used to compute the predicted state from the previous estimate, and $h$ is used to compute the predicted measurement from the predicted state. In practice, $f$ and $h$ cannot be applied directly to the covariance, and the Jacobian must be used instead. The algorithm is  shown as Algorithm~\ref{alg:alg1}.
\begin{minipage}{\textwidth}
\begin{savenotes}
\begin{algorithm}[H]
\caption{The Standard Extended Kalman filter}
\label{alg:alg1}
\begin{algorithmic}[1] 
\ENSURE for every time step $t \in T$
\STATE Let $x_{t\mid i}$ denote the estimate of state $x$ at time $t$ using observations up to time $i \leq t$ 
\STATE Let $F_{t} = \left.\dfrac{\partial f}{\partial x} \right\vert_{x_{t-1\mid t-1}, u_{t}}$ 
\STATE Let $H_{t} = \left.\dfrac{\partial f}{\partial x} \right\vert_{x_{t\mid t-1}}$
\STATE \textbf{Prediction:}
\STATE~~~~Compute predicted state estimate ${x}_{t\mid t-1} = f(x_{t-1\mid t-1}, u_{k})$
\STATE~~~~Compute predicted covariance estimate ${P}_{t\mid t-1} = F_{t}P_{t-1\mid t-1}F_{t}^{\top} + Q_{t}$
\STATE \textbf{Update:}
\STATE~~~~Compute measurement residual $y_{t} = z_{t} - h(x_{t\mid t-1})$
\STATE~~~~Compute covariance residual $S_{t} = H_{t}P_{t\mid t-1}H_{t}^{\top} + R_{t}$
\STATE~~~~Compute Kalman gain $K_{t} = P_{t\mid t-1}H_{t}^{\top}S_{t}^{-1}$
\STATE~~~~Update  state estimate $x_{t\mid t} = x_{t\mid t-1} + K_{t}y_{t}$
\STATE~~~~Update  covariance estimate $P_{t\mid t} = (I - K_{t}H_{t})P_{t\mid t-1}$
\end{algorithmic}
\end{algorithm}
\end{savenotes}
\end{minipage}

\paragraph{Extended Kalman filtering and connection problems}

Given the nature of the scenario analyzed in this paper (when contact is lost between satellite and ground station) extended Kalman filtering cannot work as intended. In this scenario, one has only two choices: to use the last received GPS position vector (we will denote this as EKFFG as described in Algorithm \ref{alg:alg2}) or  not to use GPS information in the Kalman update step at all (we will denote this as CP as described in Algorithm \ref{alg:alg3}). The second choice will seem quite familiar to those well acquainted with orbital mechanics, since this algorithm essentially solves the differential equations describing  satellite motion using Cowell's approach through Runge--Kutta methods \footnotemark[1]{} \cite{cowell1}. We implement both approaches in the scenario where the satellite has lost connection and demonstrate that no matter what choice is made, the classical approach (using the Gibbs method in tandem with extended Kalman filtering) performs significantly worse than ResNet50 and slightly worse than many CNNs. 

\footnotetext[1]{The standard definition of a Cowell propagator does not require that account be taken of drag forces \cite{CowellProp23}. In our implementation, we do account for drag, but, if desired, it can be ignored, albeit at the expense of decreased accuracy.}

\newpage

\begin{minipage}{\textwidth}
\begin{savenotes}
\begin{algorithm}[H]
\caption{Extended Kalman filtering with fixed GPS coordinates (EKFFG)}
\label{alg:alg2}
\begin{algorithmic}[1]
\ENSURE for every time step $t \in T$:
\STATE Let $x_{t\mid i}$ denote the estimate of state $x$ at time $t$ using observations up to time $i \leq t$
\STATE Let $F_{t} = \left.\dfrac{\partial f}{\partial x} \right\vert_{x_{t-1\mid t-1}, u_{t}}$
\STATE Let $H_{t} = \left.\dfrac{\partial f}{\partial x} \right\vert_{x_{t\mid t-1}}$
\STATE Let $z$ be fixed as the last received GPS position vector
\STATE \textbf{Prediction:}
\STATE~~~~Compute predicted state estimate ${x}_{t\mid t-1} = f(x_{t-1\mid t-1}, u_{k})$
\STATE~~~~Compute predicted covariance estimate ${P}_{t\mid t-1} = F_{t}P_{t-1\mid t-1}F_{t}^{\top} + Q_{t}$
\STATE \textbf{Update:}
\STATE~~~~Compute measurement residual $y_{t} = z - h(x_{t\mid t-1})$
\STATE~~~~Compute covariance residual $S_{t} = H_{t}P_{t\mid t-1}H_{t}^{\top} + R_{t}$
\STATE~~~~Compute Kalman gain $K_{t} = P_{t\mid t-1}H_{t}^{\top}S_{t}^{-1}$
\STATE~~~~Update  state estimate $x_{t\mid t} = x_{t\mid t-1} + K_{t}y_{t}$
\STATE~~~~Update  covariance estimate $P_{t\mid t} = (I - K_{t}H_{t})P_{t\mid t-1}$
\end{algorithmic}
\end{algorithm}
\end{savenotes}
\end{minipage}

\begin{minipage}{\textwidth}
\begin{savenotes}
\begin{algorithm}[H]
\caption{Cowell propagator (CP)}
\label{alg:alg3}
\begin{algorithmic}[1]
\ENSURE for every time step $t \in T$:
\STATE \textbf{Prediction:}
\STATE~~~~Compute predicted state estimate ${x}_{t\mid t-1} = f(x_{t-1\mid t-1}, u_{k})$ via Cowell's method
\end{algorithmic}
\end{algorithm}
\end{savenotes}
\end{minipage}

\subsection{Experiment and Testing}
We process the Landsat data as described in sections 3.1.2 and 3.1.3, utilizing the two datasets $\mathcal{D}_{nn}$ and $\mathcal{D}_{kf}$ we wish to determine if there is in fact a difference in performance between Neural Networks and Kalman filters when loss of connection occurs. We take the Null Hypothesis be that the resulting RMSE obtained by Kalman filtering and the Neural Network is equivalent. The Alternative hypothesis is then that the Neural Network has a lower RMSE than the Kalman filter. We observe for every Neural Network tested against both Kalman filtering approaches $p$-values $< 0.01$ indicating statistical significance and strong evidence against the null hypothesis. It is important to note that we wish to ensure independence of samples and partition our datasets in such a way that the resulting set of satellite data samples are pairwise disjoint. We utilize the paired t-test to compare the matched groups $\mathcal{P}_{nn}$ and $\mathcal{P}_{kf}$, which are the errors made by the Kalman filter and those made by the Neural Network. We get a matched sample (errors) which result from an unpaired sample. 
Additionally we conduct an F-test for equality of variance. Since we received $p$-values $<< 0.05$ for each matched group, as shown in Table \ref{table2}, we determine that the variances are not homogeneous, thereby suggesting heteroscedasticity. Since we observe that the spread is close to proportional to the mean we use $\log$ as a variance-stabilizing transformation. We construct a $95\%$ confidence interval for the errors and plot the distribution. The exact experimental algorithm is laid out below (Algorithm \ref{alg:alg4}).

\begin{table}[H]
\caption{We provide the $p$-values from the F-test justifying heteroscedasticity}
\label{table2}
\begin{tabularx}{\textwidth}{@{} l *{10}{c} c @{}}
\toprule
Comparing & $p$-value from F-test  & Is Heteroscedastic \\
\midrule
ResNet-101 vs CP &  $ 1.578488 \times 10^{-12}$ & True \\
ResNet-101 vs EKFFG & $8.101733 \times 10^{-3}$ & True \\
ResNet-50 vs CP &  $6.429322 \times 10^{-16}$  & True \\
ResNet-50 vs EKFFG &  $6.301822 \times 10^{-3}$ & True \\
AlexNet vs CP & $2.320262 \times 10^{-13}$  & True\\
AlexNet vs EKFFG & $1.028544 \times 10^{-2}$ & True\\
\bottomrule
\end{tabularx}
\end{table}

\renewcommand*\footnoterule{}
\renewcommand{\footnotesize}{\fontsize{11pt}{12pt}\selectfont}

\begin{algorithm}[H]
    \caption{Experimental Procedure}
    \label{alg:alg4}
    \textbf{Input:} A specific neural network and Kalman filtering algorithm (either Algorithm \ref{alg:alg2} or \ref{alg:alg3})\\
    \textbf{Output:} $p$-value, $t$-statistic, confidence intervals, plot of distribution
    \begin{algorithmic}[1]
        \STATE Let $\mathcal{N}$ be a neural network, and let $\mathcal{K}$ be a Kalman filter
        \STATE Let $N$ be the number of trials, fix $N = 30$, and let $\mathcal{I}$ be an index set $\{0, \ldots, 29\}$
        \STATE Let $\mathcal{P}_{nn} \leftarrow Array()$ be an initially empty array recording $\mathcal{N}$'s performance
        \STATE Let $\mathcal{P}_{kf} \leftarrow Array()$ be an initially empty array recording $\mathcal{K}$'s performance
        \STATE Split $\mathcal{D}_{nn}$ into $(\text{train}_{nn}, \text{validation}_{nn}, \text{test}_{nn})$
        \STATE Partition: 
        \STATE~~~~$\text{train}_{nn}$ into $N$ many disjoint subsets
        \STATE~~~~$\text{validation}_{nn}$ into $N$ many disjoint subsets
        \STATE~~~~$\text{test}_{nn}$ into $N$ many disjoint subsets
        \STATE~~~~Yield: $\{(\text{train}_{nn,0}, \text{validation}_{nn,0}, \text{test}_{nn,0}), \ldots, (\text{train}_{nn,29}, \text{validation}_{nn,29}, \text{test}_{nn,29}) \}$
        \STATE Partition $\mathcal{D}_{kf}$ into $N$ many disjoint subsets and Yield: $\{\mathcal{D}_{kf,0}, \ldots, \mathcal{D}_{kf,29}\}$
       
        \FOR{$i \in \mathcal{I}$}
            \STATE Select/Access the $i$-th partition $(\text{train}_{nn,i}, \text{validation}_{nn,i}, \text{test}_{nn,i})$
            \STATE Initialize $\mathcal{N}$ with the ImageNet weights
            \STATE Train $\mathcal{N}$ using $\text{train}_{nn,i}$ with SGD and use $\text{validation}_{nn,i}$ for hyperparameter tuning
            \STATE Evaluate the test-set RMSE $r_{nn}$ for $\text{test}_{nn,i}$
            \STATE Append$(\mathcal{P}_{nn}, r_{nn})$
            \STATE Select/Access the $i$-th partition of $\mathcal{D}_{kf}$, in essence $\mathcal{D}_{kf,i}$
            \STATE Initialize the Kalman filter object $\mathcal{K}$ with the data from Gibbs Method
            \STATE Get the predicted estimate for $\mathcal{D}_{kf,i}$ from $\mathcal{K}$
            \STATE Evaluate the RMSE $r_{kf}$ between the predicted estimate and true value
            \STATE Append$(\mathcal{P}_{kf}, r_{kf})$
        \ENDFOR
        \STATE Using the two lists of error values $P_{nn}$ and $P_{kf}$ we gather the following statistics:
        \STATE~~~~Test for heteroscedasticity (F-test, Scipy) $\rightarrow$ apply $\log$ transformation if $p < 0.05$
        \STATE~~~~Compute Pearson correlation using Scipy
        \STATE~~~~Conduct a paired sample t-test using Scipy $\rightarrow$ ($p$-value, $t$-statistic)
        \STATE~~~~Plot the error distribution\
        \STATE~~~~Compute $95\%$ confidence intervals for both $P_{nn}$ and $P_{kf}$
    \end{algorithmic}
\end{algorithm}

\newpage

\section{Results}
We conduct the above experiment as described by running Algorithm \ref{alg:alg4}.
We compare every Neural Network: ResNet101, ResNet50, and AlexNet with both Kalman filters: EKFFG (Algorithm \ref{alg:alg2}) and CP (Algorithm \ref{alg:alg3}).

\subsubsection{Hypothesis Test}
\noindent
(Null) $H_{0}$: $\mu_{NN} = \mu_{KF}$\\
(Alternative) $H_{A}$: $\mu_{NN} < \mu_{KF}$\\
Where $\mu_{NN}$ is the mean test-set RMSE for the Neural Network and $\mu_{KF}$ is the mean evaluation RMSE for the Kalman filter.\\
More verbosely, we let the null hypothesis be that $\mu_{NN}$ equals the mean evaluation RMSE for the Kalman filter $\mu_{KF}$. Correspondingly, we let the alternative hypothesis be that the mean test-set RMSE for the Neural Network $\mu_{NN}$ is less than the mean evaluation RMSE for the Kalman filter $\mu_{KF}$.\\

\subsubsection{Neural Network and Kalman filter Comparison}
\noindent
We plot the error distributions comparing the performances of ResNet-101 with the Cowell Propagator and EKFFG in Figures \ref{fig: fig1} and \ref{fig: fig2}, ResNet50 with the Cowell Propagator and EKFFG in Figures \ref{fig: fig3} and \ref{fig: fig4}, AlexNet with the Cowell Propagator and EKFFG in Figures \ref{fig: fig5} and \ref{fig: fig6}. 
In each case, we observe statistically significant results ($p$ value $<< 0.01$) and strong evidence against the null hypothesis, leading us to accept the alternative hypothesis and conclude the Neural Networks outperform the Kalman filtering approaches. We summarize the results of our experiments comparing the performance of the Neural Networks to that of the Kalman filtering approaches in Table \ref{table3}.

\begin{table}[H]
\caption{We summarize our experiments comparing Neural Networks to Kalman filtering}
\label{table3}
\begin{tabularx}{\textwidth}{@{} l *{10}{c} c @{}}
\toprule
Comparing & $p$-value  & $t$-statistic  & Neural Net Confidence Interval & Kalman filter Confidence Interval \\
\midrule
ResNet-101 vs CP &  $ 4.941977 \times 10^{-9}$ & $-7.917213674$ & $(4.51847, 4.97118) $ & $(5.79643, 6.11472)$\\
ResNet-101 vs EKFFG & $6.380400 \times 10^{-11}$ & $-9.711756015$ & $(4.41143, 4.79816)$ & $(6.11150, 6.85394)$\\
ResNet-50 vs CP &  $1.343550 \times 10^{-11}$  & $-10.400605650$ & $(4.24241, 4.62236)$ & $(5.79643, 6.11472)$\\
ResNet-50 vs EKFFG &  $2.107189 \times 10^{-12}$ & $-11.256603426$ & $(4.02317, 4.35008)$ & $(6.11150, 6.85394)$\\
AlexNet vs CP & $2.17117 \times 10^{-11}$  & $-10.185497814$ & $(5.03809, 5.1161)$& $(5.79643, 6.11472)$ \\
AlexNet vs EKFFG & $3.056926 \times 10^{-9}$ &  $-8.010718261$ &  $(5.01128, 5.07986)$ & $(6.11150, 6.85394)$\\
\bottomrule
\end{tabularx}
\end{table}

\newpage

\section{Discussion}

We find that in the event that a satellite has lost connection with its ground-based station, a computer vision based approach, relying on CNNs and satellite images of the earth, is able to predict the satellite's orbit better than the EKFFG (Algorithm \ref{alg:alg2}) and CP (Algorithm \ref{alg:alg3}). We thus show that machine learning can be used for full orbit determination; essentially, CNNs are able to completely replace standard state propagators, relying on minimal hardware (a RISC-V processor and camera). We further address the limitations posed by other deep learning approaches (AML-EP, AML-ERP, RKF-EP) for satellite orbit prediction, as discussed in section 2.3.

We see numerous directions for future work. While we demonstrate the effectiveness of our approach, the incorporation of continuous GNSS measurements may aid the CNNs' prediction capabilities. We would also like to further experiment with the deployment of these CNNs to resource limited devices through quantization and model pruning, which could lead to an extremely computationally efficient approach to performing orbit determination.

\section*{Acknowledgment}

We would like to acknowledge Google for making available Landsat Data. We would also like to acknowledge Nilesh Chaturvedi, Arya Das, and Alexandros Kazantzidis for making their Gibbs method and Kalman filter code publicly available. 

\bibliographystyle{unsrt}  
\bibliography{references}  

\section*{Figure captions}
We provide the figures for the results section below.

\begin{center}
\begin{minipage}{0.48\linewidth}
\begin{figure}[H]
    \includegraphics[width=\linewidth]{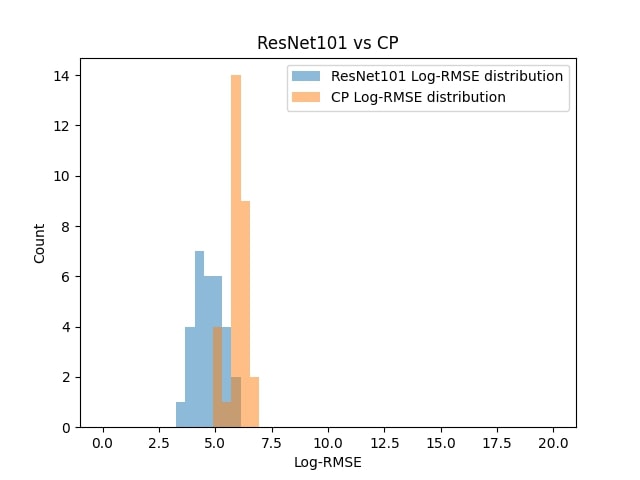}
    \caption{Log RMSE of ResNet-101 vs CP}
    \label{fig: fig1}
\end{figure}
\end{minipage}
\hfill
\begin{minipage}{0.49\linewidth}
\begin{figure}[H]
    \includegraphics[width=\linewidth]{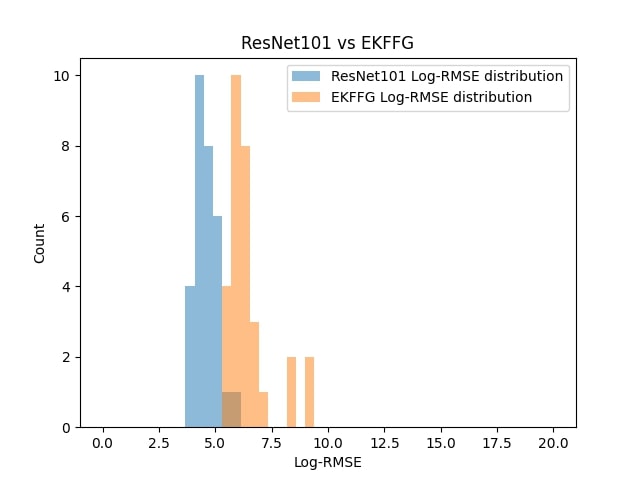}
    \caption{Log RMSE of ResNet-101 vs EKFFG}
    \label{fig: fig2}
\end{figure}
\end{minipage}
\end{center}

\begin{center}
\begin{minipage}{0.49\linewidth}
\begin{figure}[H]
    \includegraphics[width=\linewidth]{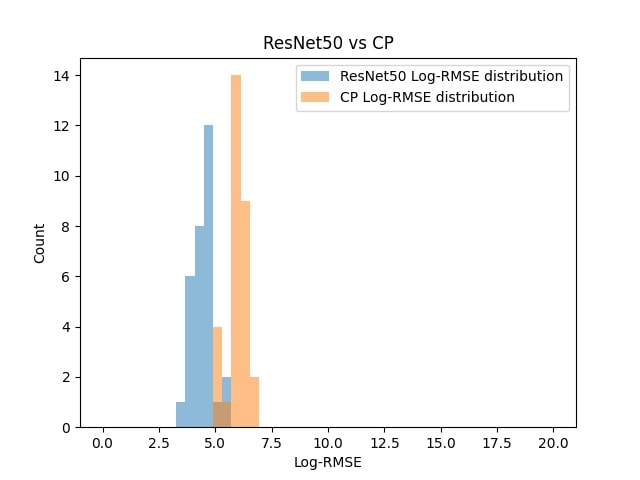}
    \caption{Log RMSE of ResNet-50 vs CP}
    \label{fig: fig3}
\end{figure}
\end{minipage}
\hfill
\begin{minipage}{0.49\linewidth}
\begin{figure}[H]
    \includegraphics[width=\linewidth]{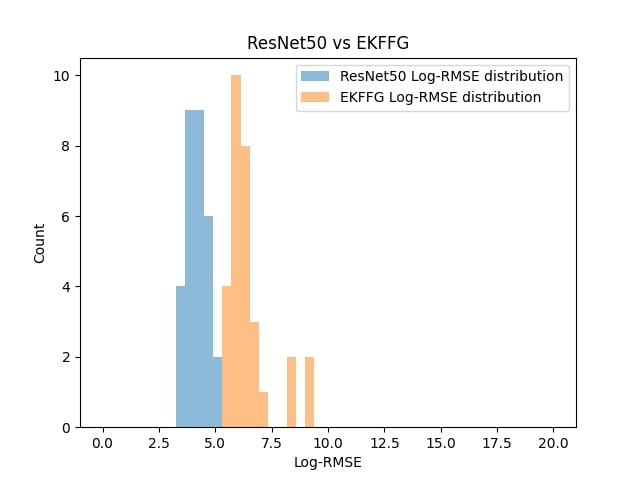}
    \caption{Log RMSE of ResNet-50 vs EKFFG}
    \label{fig: fig4}
\end{figure}
\end{minipage}
\end{center}

\begin{center}
\begin{minipage}{0.49\linewidth}
\begin{figure}[H]
    \includegraphics[width=\linewidth]{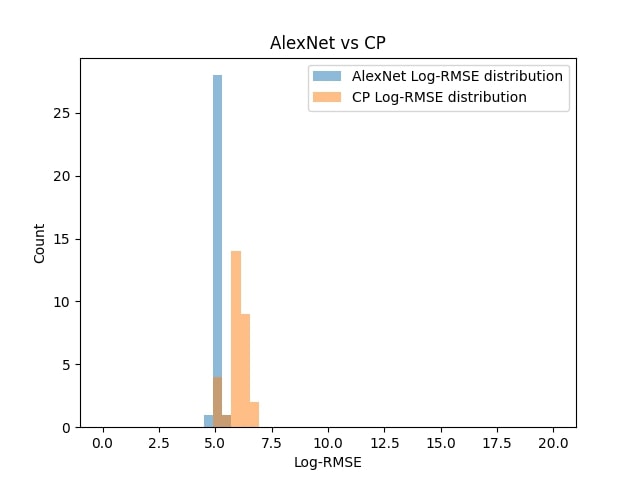}
    \caption{Log RMSE of AlexNet vs CP}
    \label{fig: fig5}
\end{figure}
\end{minipage}
\hfill
\begin{minipage}{0.49\linewidth}
\begin{figure}[H]
    \includegraphics[width=\linewidth]{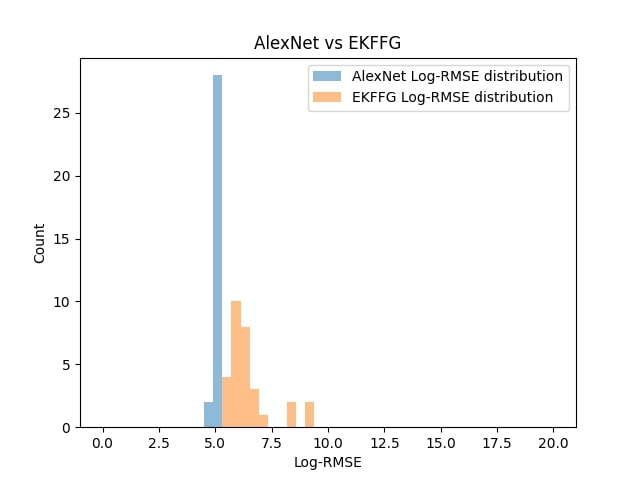}
    \caption{Log RMSE of AlexNet vs EKFFG}
    \label{fig: fig6}
\end{figure}
\end{minipage}
\end{center}

\end{document}